# Recognizing Combinations of Facial Action Units with Different Intensity Using a Mixture of Hidden Markov Models and Neural Network

Mahmoud Khademi, Mohammad Taghi Manzuri-Shalmani,
Mohammad Hadi Kiapour, and Ali Akbar Kiaei

DSP Lab, Sharif University of Technology, Tehran, Iran
{khademi@ce.,manzuri@,kiapour@ee.,kiaei@ce.}sharif.edu

**Abstract.** Facial Action Coding System consists of 44 action units (AUs) and more than 7000 combinations. Hidden Markov models (HMMs) classifier has been used successfully to recognize facial action units (AUs) and expressions due to its ability to deal with AU dynamics. However, a separate HMM is necessary for each single AU and each AU combination. Since combinations of AU numbering in thousands, a more efficient method will be needed. In this paper an accurate real-time sequence-based system for representation and recognition of facial AUs is presented. Our system has the following characteristics: 1) employing a mixture of HMMs and neural network, we develop a novel accurate classifier, which can deal with AU dynamics, recognize subtle changes, and it is also robust to intensity variations, 2) although we use an HMM for each single AU only, by employing a neural network we can recognize each single and combination AU, and 3) using both geometric and appearance-based features, and applying efficient dimension reduction techniques, our system is robust to illumination changes and it can represent the temporal information involved in formation of the facial expressions. Extensive experiments on Cohn-Kanade database show the superiority of the proposed method, in comparison with other classifiers.

## 1 Introduction

Human face-to-face communication is a standard of perfection for developing a natural, robust, effective and flexible multi modal/media human-computer interface due to multimodality and multiplicity of its communication channels. In this type of communication, the facial expressions constitute the main modality [1]. In this regard, automatic facial expression analysis can use the facial signals as a new modality and it causes the interaction between human and computer more robust and flexible. Moreover, automatic facial expression analysis can be used in other areas such as lie detection, neurology, intelligent environments and clinical psychology.

Facial expression analysis includes both measurement of facial motion (e.g. mouth stretch or outer brow raiser) and recognition of expression (e.g. surprise or anger). Real-time fully automatic facial expression analysis is a challenging complex topic in computer vision due to pose variations, illumination variations, different age, gender, ethnicity, facial hair, occlusion, head motions, and lower intensity of expressions. Two survey papers summarized the work of facial expression analysis before year 1999 [2, 3]. Regardless of the face detection stage, a typical automatic facial expression analysis system consists of facial expression data extraction and facial expression classification stages. Facial feature processing may happen either holistically, where the face is processed as a whole, or locally. Holistic feature extraction methods are good at determining prevalent facial expressions, whereas local methods are able to detect subtle changes in small areas.

There are mainly two approaches for facial data extraction: geometric-based methods and appearance-based methods. The geometric facial features present the shape and locations of facial components. With appearance-based methods, image filters, e.g. Gabor wavelets, are applied to either the whole face or specific regions in a face image to extract a feature vector [4].

The sequence-based recognition method uses the temporal information of the sequences (typically from natural face towards the frame with maximum intensity) to recognize the expressions. To use the temporal information, the techniques such as hidden Markov models (HMMs) [5], recurrent neural networks [6] and rule-based classifier [7] were applied.

The facial action coding system (FACS) is a system developed by Ekman and Friesen [8] to detect subtle changes in facial features. The FACS is composed of 44 facial action units (AUs). 30 AUs of them are related to movement of a specific set of facial muscles: 12 for upper face and 18 for lower face (see Table 1).

**Table 1.** Some upper face and lower face action units (for more details see [8, 9])

| Upper face AU | Description | Lower face AU | Description |
| --- | --- | --- | --- |
| AU 1 | Inner brow raiser | AU 12 | Lip corner puller |
| AU 2 | Outer brow raiser | AU 15 | Lip corner depressor |
| AU 4 | Brow lowerer | AU 17 | Chin raiser |
| AU 5 | Upper lid raiser | AU 20 | Lip stretcher |
| AU 6 | Cheek raiser | AU 23 | Lip tightener |
| AU 7 | Lid tightener | AU 24 | Lip pressor |
| Lower face AU | Description | AU 25 | Lip parts |
| AU 9 | Nose wrinkle | AU 26 | Jaw drop |
| AU 10 | Upper lip raiser | AU 27 | Mouth stretch |

Facial action units can occur in combinations and vary in intensity. Although the number of single action units is relatively small, more than 7000 different AU combinations have been observed. They may be *additive*, in which the combination does not change the appearance of the constituent single AUs, or *nonadditive*, in which the appearance of the constituent single AUs does change. To capture such subtlety of human emotion paralinguistic communication, automated recognition of fine-grained changes in facial expression is required (for more details see [8, 9]).

In this paper an accurate real-time sequence-based system for representation and recognition of facial action units is presented. We summarize the advantages of our system as follows:

1) We developed a classification scheme based on a mixture of HMMs and neural network, which can deal with AU dynamics, recognize subtle changes, and it is also robust to intensity variations.

2) HMMs classifier can deal with AU dynamics properly. But, it is impossible to use a separate HMM for each AU combination since the combinations numbering in thousands. We use an HMM for each single AU only. However, by employing a neural network we can recognize each single AU and each combination AU. Also, we use an accurate method for training the HMMs by considering the intensity of AUs.

3) Recent work suggests that spontaneous and deliberate facial expressions may be discriminated in term of timing parameters. Employing temporal information instead of using only the last frame, we can represent these parameters properly. Also, using both geometric and appearance features, we can increase the recognition rate and make the system robust against illumination changes.

4) By employing rule-based classifiers, we can automatically extract human interpretable classification rules to interpret each expression using continues values of AU intensity.

5) Due to the relatively low computational cost in the test phase, the proposed system is suitable for real-time applications.

The rest of the paper has been organized as follows: In section 2, we describe the approach which is used for facial data extraction and representation using both geometric and appearance features. Then, we discuss the proposed scheme for recognition of facial action units in section 3. Section 4 reports our experimental results, and section 5 presents conclusions and a discussion.

## 2  Facial Data Extraction and Representation

### 2.1  Geometric-Based Facial Feature Extraction Using Optical Flow

In order to extract geometric features, the points of a 113-point grid, which is called Wincandide-3, are placed on the first frame manually. Automatic registering of the grid with the face has been addressed in many literatures (e.g. see [10]). For upper face and lower face action units a particular subset of points are selected (see Fig. 1a). The pyramidal optical flow tracker [11] is employed to track the points of the model in the successive frames towards the last frame (see Fig. 1b). The loss of the tracked points is handled through a model deformation procedure (for details see [12]). For each frame, the displacements of the points in two directions with respect to the first frame are calculated to form a feature vector. Assume the last frames of the sequences in the training set, have the maximum intensity. Also, define:

$$\text{intensity}(f) = \frac{\text{sumdistances}_1(f)}{\text{sumdistances}_2} \tag{1}$$

where $\text{sumdistances}_1$ is the sum of the Euclidian distances between point of the Wincandide-3 grid in fth frame of the sequence and their positions in the first frame (a subset of points for upper face and lower face action units are used). Similarly, $\text{sumdistances}_2$ is the sum of the Euclidean distances between points of the model in the last frame of the sequence and their positions in the first frame; e.g. if the sequence contains t frames, then intensity(t) = 1, and intensity(1) = 0. The frames of each sequence in the training set are divided into three subsets, i.e. three states, based on their intensity:

$0 \leq$ intensity(f) $< 0.33$,  $0.33 \leq$ intensity(f) $\leq 0.66$, and $0.66 <$ intensity(f) $\leq 1$

(For some of the single AUs we use five subsets (states)). Then, we apply principal component analysis (PCA) algorithm [13], separately to the feature vectors of these subsets to form the final geometric feature vectors of each state.

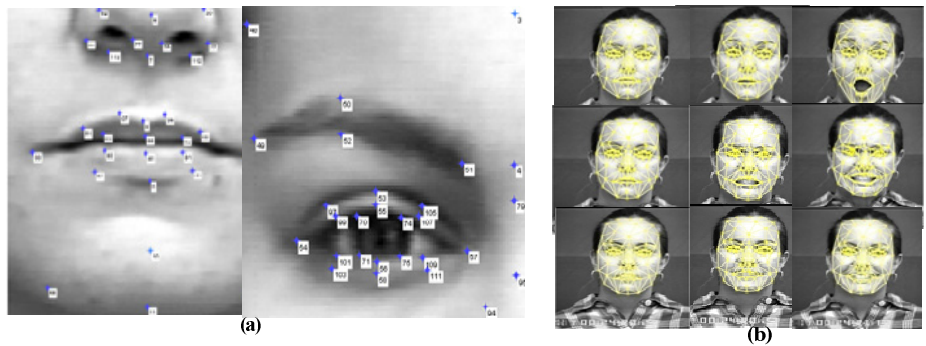

**Fig. 1.** a. Selected points for upper face and lower face action units. b. Geometric-based facial feature extraction using feature point tracking.

## 2.2  Appearance-Based Facial Feature Extraction Using Gabor Wavelets

In order to extract the appearance-based facial features from each frame, we use a set of Gabor wavelets. They allow detecting line endings and edge borders of each frame over multiple scales and with different orientations. Gabor wavelets remove also most of the variability in images that occur due to lighting changes [4]. Each frame is convolved with p wavelets to form the Gabor representation of the t frames (Fig. 2).

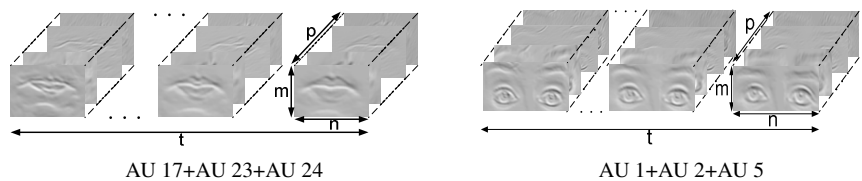

**Fig. 2.** Examples of the image sequences and their representation using Gabor wavelets

The frames of each sequence in the training set are divided into three or five subsets (states), based on grouping that we discussed in the previous subsection. In order to embed facial features in a low-dimensionality space and deal with curse of dimensionality dilemma, we should use a dimension reduction method. We apply 2D principal component analysis (2DPCA) algorithm [14] to each feature matrices of each subset separately. Then, we concatenate the vectorized representation of the reduced feature matrices of each frame, and apply a 1DPCA [13] to them. The resulted feature vectors from each frame, are used for classification.

## 3   Facial Action Unit Recognition

The flow diagram of the proposed system is shown in Fig. 3. We will assume that sth $(s = 1,2,...,M)$ HMM is characterized by the following set of parameters (M is the number of HMMs, i.e. the number of single AUs):

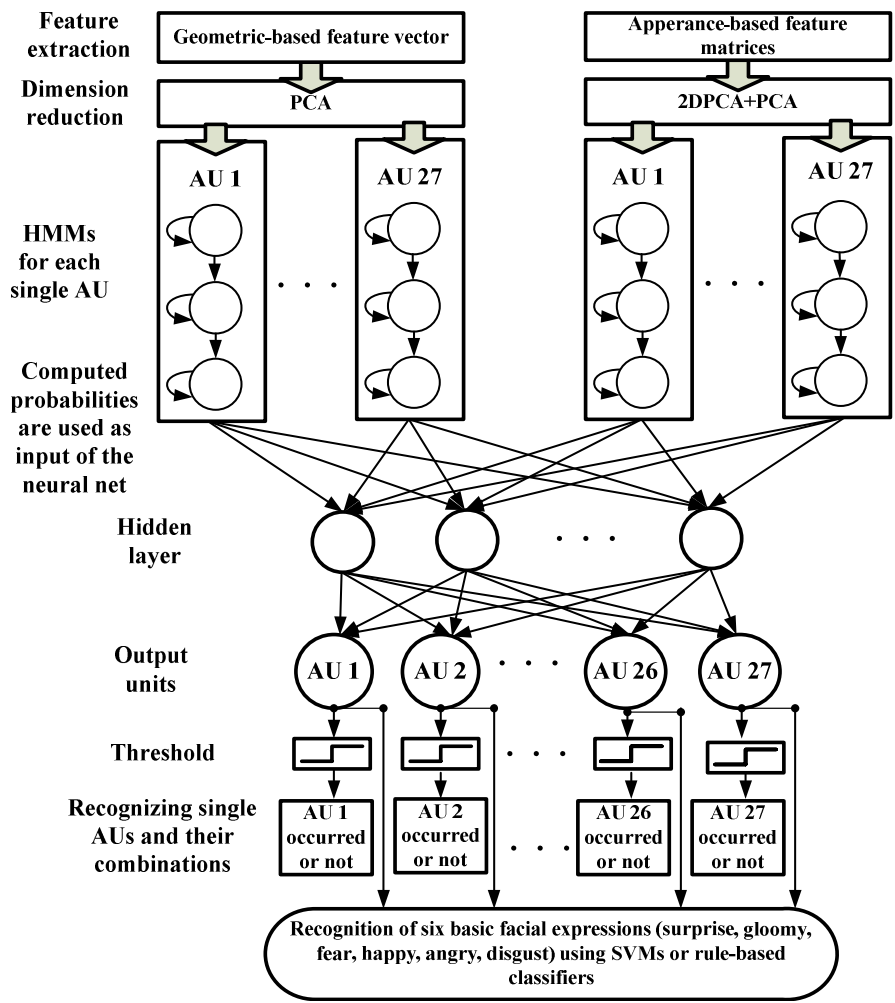

**Fig. 3.** Flow diagram of the proposed system

1) $N_s$, the number of the states of sth HMM (we use three or five-state left to right HMM for each single AU).
2) The probability densities $p(x|j)$ $(j = 1,2,\ldots,N_s)$, describing the distribution of the observations emitted from state j (we use three Gaussians for each state, the mean and variance of the Gaussians are computed using training data and maximum likelihood estimation method).
3) The transition probabilities $P(i|j)$ $(i,j = 1,2,\ldots,N_s)$, among the various states (the transition probabilities, which have not been depicted in left to right HMMs of Fig. 3, are zero, for others we use equal transition probabilities).
4) The probabilities $P(i)$ $(i = 1,2,\ldots,N_s)$, of the initial state $(P(1) = 1$ and $P(j) = 0$ for $j \neq 1$ because each sequence is started from natural face).
The training phase contains two steps: In the first step, we find probability density functions. Then, having observed sequence of feature vectors $X: x_1,\ldots,x_t$, (t is number of frames) in the respective sequence of states $\Omega_i^s: \omega_{i_1}^s, \ldots, \omega_{i_t}^s$, the quantity:

$$P(X|\Omega_*^s) = \max_i P(X|\Omega_i^s) = \max_i P(\omega_{i_1}^s)p(x_1|\omega_{i_1}^s)\prod_{k=2}^{N_s} P(\omega_{i_k}^s|\omega_{i_{k-1}}^s)p(x_k|\omega_{i_k}^s) \quad (2)$$

is computed for each of the M reference models. Its efficient computation can be achieved via Viterbi algorithm. In the second step of the training phase, we train the neural network by the values $P(X|\Omega_*^s)$ $(s = 1,\ldots,M)$ as the inputs and using back propagation algorithm. We trained the neural network with an output unit for each single AU and by allowing multiple output units to fire when the input sequence consists of AU combinations. The sth value of the target vector is 1 if sth single AU occurs in the corresponding image sequence, otherwise it would be 0). Moreover, we can use continuous values between 0 and 1 for sequences with different intensity of expression, as target value of the feature vectors. We also use some sequences several times with different intensities, i.e. by using intermediate frames as the last frame and removing the frames which come after it. Applying this method, we can properly recognize lower intensity and combinations of AUs. We model each single AU two times, using geometric and appearance features separately. In test phase, having observed sequence of feature vectors $X: x_1,\ldots,x_t$, in the sequence of states $\Omega_i^s: \omega_{i_1}^s,\ldots,\omega_{i_t}^s$, the quantities $P(X|\Omega_*^s)$ $(s = 1,\ldots,M)$ are first computed. Then, the outputs of the neural network are passed through a threshold (we set it 0.5). When several outputs are on, it signals that a combination of AUs has been occurred.

Although we can use a SVMs for classification of six basic facial expressions (by feature vectors directly or AU intensity values), employing rule-based classifiers such as JRIP [15], we can automatically extract human interpretable classification rules to interpret each expression. Thus, novel accurate AU-to-expression converters by continues values of the AU intensity can be created. These converters would be useful in animation, cognitive sciences, and behavioral sciences areas.

## 4   Experimental Results

To evaluate the performance of the proposed system and other methods like support vector machines (SVMs) [12], mixture of HMMs and SVMs, and neural network (NN) classifiers, we test them on Cohn-Kanade database [16]. The database includes 490 frontal view image sequences from over 97 subjects. The final frame of each image sequence has been coded using Facial Action Coding System which describes subject's expression in terms of action units. For action units that vary in intensity, a 5-point ordinal scale has been used to measure the degree of muscle contraction. In order to test the algorithm in lower intensity situation, we used each sequence five times with different intensities, i.e. by using intermediate frames as the last frame. Of theses, 1500 sequences were used as the training set. Also, for upper face and lower face AUs, 240 and 280 sequences were used as the test set respectively. None of the test subjects appeared in training data set. Some of the sequences contained limited head motion.

Image sequences from neutral towards the frame with maximum intensity, were cropped into $57 \times 102$ and $52 \times 157$ pixel arrays for lower face and upper face action units respectively. To extract appearance features we applied 16 Gabor kernels to each frame. After applying the dimension reduction techniques, depending on the single AU that we want to model it, the geometric and appearance feature vectors were of dimension 6 to 10 and 40 to 60 respectively. The best performance was obtained by three and five states HMMs depend on the corresponding single AU. Table 2 and Table 3 show the upper face and lower face action unit recognition results respectively. In the proposed method, an average recognition rate of 90.0 and 96.1 percent were achieved for upper face and lower face action units respectively. Also, an average false alarm rate of 5.8 and 2.5 percent were achieved for upper face and lower face action units respectively.

**Table 2.** Upper face action unit recognition results (R=recognition rate, F=false alarm)

| Proposed method (HMMs+NN) | | | | | HMMs+SVMs | | | | |
|---|---|---|---|---|---|---|---|---|---|
| AUs | Sequences | Recognized AUs | | | AUs | Sequences | Recognized AUs | | |
| | | True | Missing or extra | False | | | True | Missing or extra | False |
| 1 | 20 | 18 | 1(1+2+4), 1(1+2) | 0 | 1 | 20 | 17 | 2(1+2+4), 1(1+2) | 0 |
| 2 | 10 | 8 | 2(1+2) | 0 | 2 | 10 | 6 | 2(1+2+4), 1(1+2) | 1(1) |
| 4 | 20 | 19 | 1(1+2+4) | 0 | 4 | 20 | 18 | 1(1+2+4) | 1(2) |
| 5 | 20 | 20 | 0 | 0 | 5 | 20 | 20 | 0 | 0 |
| 6 | 20 | 19 | 0 | 1(7) | 6 | 20 | 18 | 1(1+6) | 1(7) |
| 7 | 10 | 9 | 0 | 1(6) | 7 | 10 | 7 | 3(6+7) | 0 |
| 1+2 | 40 | 37 | 2(2), 1(1+2+4) | 0 | 1+2 | 40 | 38 | 1(1+2+4) | 1(4) |
| 1+2+4 | 20 | 18 | 1(1), 1(2) | 0 | 1+2+4 | 20 | 18 | 1(2), 1(1+2) | 0 |
| 1+2+5 | 10 | 8 | 2(1+2) | 0 | 1+2+5 | 10 | 7 | 3(1+2) | 0 |
| 1+4 | 10 | 8 | 2(1+2+4) | 0 | 1+4 | 10 | 5 | 3(1+2+4) | 2(5) |
| 1+6 | 10 | 8 | 1(1+6+7) | 1(7) | 1+6 | 10 | 6 | 2(1+6+7) | 2(7) |
| 4+5 | 20 | 18 | 1(4), 1(5) | 0 | 4+5 | 20 | 15 | 2(4), 1(5) | 2(2) |
| 6+7 | 30 | 27 | 2(1+6+7), 1(7) | 0 | 6+7 | 30 | 25 | 3(1+6+7), 2(7) | 0 |
| Total | 240 | 217 | 20 | 3 | Total | 240 | 200 | 30 | 10 |
| R | 90.0% | | | | R | 83.3% | | | |
| F | 5.8% | | | | F | 12.5% | | | |

| SVMs | | | | | NN [17] | | | | |
|---|---|---|---|---|---|---|---|---|---|
| AUs | Sequences | Recognized AUs | | | AUs | Sequences | Recognized AUs | | |
| | | True | Missing or Extra | False | | | True | Missing or Extra | False |
| 1 | 20 | 15 | 2(1+2+4), 1(1+2) | 2(2) | 1 | 20 | 14 | 3(1+2+4) | 3(2) |
| 2 | 10 | 6 | 2(1+2+4) | 2(1) | 2 | 10 | 5 | 4(1+2+4) | 1(1) |
| 4 | 20 | 18 | 1(1+2+4) | 1(2) | 4 | 20 | 18 | 1(1+2+4) | 1(2) |
| 5 | 20 | 20 | 0 | 0 | 5 | 20 | 18 | 1(4+5) | 1(5) |
| 6 | 20 | 19 | 1(1+6) | 0 | 6 | 20 | 18 | 2(1+6) | 0 |
| 7 | 10 | 7 | 0 | 3(6) | 7 | 10 | 6 | 2(6+7) | 2(6) |
| 1+2 | 40 | 35 | 1(2), 2(1+2+4) | 2(4) | 1+2 | 40 | 36 | 2(2), 2(1+2+4) | 0 |
| 1+2+4 | 20 | 15 | 2(1), 2(2) | 1(5) | 1+2+4 | 20 | 16 | 2(1), 2(2) | 0 |
| 1+2+5 | 10 | 6 | 2(1+5) | 2(4) | 1+2+5 | 10 | 7 | 2(1+2) | 1(4) |
| 1+4 | 10 | 4 | 3(1+2+4) | 3(5) | 1+4 | 10 | 5 | 4(1+2+4) | 1(5) |
| 1+6 | 10 | 6 | 3(1+6+7) | 1(7) | 1+6 | 10 | 6 | 2(1+6+7) | 2(7) |
| 4+5 | 20 | 15 | 2(1+2+5) | 3(2) | 4+5 | 20 | 16 | 3(4) | 1(1) |
| 6+7 | 30 | 24 | 2(1+6+7), 2(7) | 2(1) | 6+7 | 30 | 24 | 3(1+6+7), 3(7) | 0 |
| Total | 240 | 190 | 28 | 22 | Total | 240 | 189 | 38 | 13 |
| R | 79.2% | | | | R | 78.8% | | | |
| F | 17.1% | | | | F | 15.4% | | | |

In HMMs+SVMs method, for each single AU an HMM was trained. Then we classify the quantities $P(X|\Omega_*^s)$ ($s = 1, ..., M$) using M two-class (occurred or not) SMVs classifiers [12] with Gaussian kernel.

**Table 3.** Lower facial action unit recognition results (R=recognition rate, F=false alarm)

| Proposed method (HMMs+NN) | | | | | HMMs+SVMs | | | | |
|---|---|---|---|---|---|---|---|---|---|
| AUs | Sequences | Recognized AUs | | | AUs | Sequences | Recognized AUs | | |
| | | True | Missing or extra | False | | | True | Missing or extra | False |
| 9 | 8 | 8 | 0 | 0 | 9 | 8 | 8 | 0 | 0 |
| 10 | 12 | 12 | 0 | 0 | 10 | 12 | 12 | 0 | 0 |
| 12 | 12 | 12 | 0 | 0 | 12 | 12 | 12 | 0 | 0 |
| 15 | 8 | 8 | 0 | 0 | 15 | 8 | 6 | 2(15+17) | 0 |
| 17 | 16 | 16 | 0 | 0 | 17 | 16 | 16 | 0 | 0 |
| 20 | 12 | 12 | 0 | 0 | 20 | 12 | 12 | 0 | 0 |
| 25 | 48 | 48 | 0 | 0 | 25 | 48 | 45 | 1(25+26) | 2(26) |
| 26 | 24 | 18 | 4(25+26) | 2(25) | 26 | 24 | 19 | 3(25+26) | 2(25) |
| 27 | 24 | 24 | 0 | 0 | 27 | 24 | 24 | 0 | 0 |
| 9+17 | 24 | 24 | 0 | 0 | 9+17 | 24 | 22 | 2(9) | 0 |
| 9+17+23+24 | 4 | 3 | 1(19+17+24) | 0 | 9+17+23+24 | 4 | 2 | 2(19+17+24) | 0 |
| 9+25 | 4 | 4 | 0 | 0 | 9+25 | 4 | 4 | 0 | 0 |
| 10+17 | 8 | 5 | 2(17), 1(10) | 0 | 10+17 | 8 | 3 | 2(10+12) | 3(12) |
| 10+15+17 | 4 | 4 | 0 | 0 | 10+15+17 | 4 | 2 | 2(15+17) | 0 |
| 10+25 | 8 | 8 | 0 | 0 | 10+25 | 8 | 5 | 3(25) | 0 |
| 12+25 | 16 | 16 | 0 | 0 | 12+25 | 16 | 16 | 0 | 0 |
| 12+26 | 8 | 7 | 1 (12+25) | 0 | 12+26 | 8 | 5 | 2(12+25) | 1(25) |
| 15+17 | 16 | 16 | 0 | 0 | 15+17 | 16 | 16 | 0 | 0 |
| 17+23+24 | 8 | 8 | 0 | 0 | 17+23+24 | 8 | 7 | 1(17+23) | 0 |
| 20+25 | 16 | 16 | 0 | 0 | 20+25 | 16 | 13 | 3(20+26) | 0 |
| Total | 280 | 269 | 9 | 2 | Total | 280 | 249 | 23 | 8 |
| R | 96.1% | | | | R | 88.9% | | | |
| F | 2.5% | | | | F | 7.5% | | | |

| SVMs | | | | | NN [17] | | | | |
|---|---|---|---|---|---|---|---|---|---|
| AUs | Sequences | Recognized AUs | | | AUs | Sequences | Recognized AUs | | |
| | | True | Missing or extra | False | | | True | Missing or extra | False |
| 9 | 8 | 8 | 0 | 0 | 9 | 8 | 7 | 1(9+17) | 0 |
| 10 | 12 | 8 | 2(10+7) | 2(17) | 10 | 12 | 8 | 2(10+7) | 2(17) |
| 12 | 12 | 12 | 0 | 0 | 12 | 12 | 11 | 1(12+25) | 0 |
| 15 | 8 | 6 | 2(15+17) | 0 | 15 | 8 | 6 | 2(15+17) | 0 |
| 17 | 16 | 14 | 2(10+17) | 0 | 17 | 16 | 13 | 2(10+17) | 1(10) |
| 20 | 12 | 12 | 0 | 0 | 20 | 12 | 12 | 0 | 0 |
| 25 | 48 | 43 | 2(25+26) | 3(26) | 25 | 48 | 42 | 3(25+26) | 3(26) |
| 26 | 24 | 18 | 3(25+26) | 3(25) | 26 | 24 | 19 | 2(25+26) | 3(25) |
| 27 | 24 | 24 | 0 | 0 | 27 | 24 | 23 | 1(27+25) | 0 |
| 9+17 | 24 | 22 | 2(9) | 0 | 9+17 | 24 | 22 | 2(9) | 0 |
| 9+17+23+24 | 4 | 1 | 3(9+17+24) | 0 | 9+17+23+24 | 4 | 1 | 3(9+17+24) | 0 |
| 9+25 | 4 | 4 | 0 | 0 | 9+25 | 4 | 4 | 0 | 0 |
| 10+17 | 8 | 2 | 4(10+12) | 2(12) | 10+17 | 8 | 4 | 2(10+12) | 2(12) |
| 10+15+17 | 4 | 2 | 2(15+17) | 0 | 10+15+17 | 4 | 2 | 2(15+17) | 0 |
| 10+25 | 8 | 7 | 1(25) | 0 | 10+25 | 8 | 7 | 1(25) | 0 |
| 12+25 | 16 | 16 | 0 | 0 | 12+25 | 16 | 16 | 0 | 0 |
| 12+26 | 8 | 3 | 3(12+25) | 2(25) | 12+26 | 8 | 3 | 3(12+25) | 2(25) |
| 15+17 | 16 | 16 | 0 | 0 | 15+17 | 16 | 16 | 0 | 0 |
| 17+23+24 | 8 | 6 | 2(17+24) | 0 | 17+23+24 | 8 | 6 | 2(17+24) | 0 |
| 20+25 | 16 | 11 | 3(20+26) | 2(26) | 20+25 | 16 | 12 | 3(20+26) | 1(26) |
| Total | 280 | 235 | 31 | 14 | Total | 280 | 234 | 32 | 14 |
| R | 83.9% | | | | R | 83.6% | | | |
| F | 12.5% | | | | F | 12.9% | | | |

Although this method can deal with AU dynamics properly, due to use of crisp value for targets, this method suffers from intensity variations. In SVMs method, we first concatenated the reduced geometric and appearance feature vectors for each single AU. Then, we classify them using M two-class (occurred or not) SMVs classifiers with Gaussian kernel. This method cannot deal with AU dynamics. Moreover, due to use of crisp value for targets, it suffers from intensity variations. Finally, in NN methods we trained a neural network (NN) with an output unit for each single AU and by allowing multiple output units to fire when the input sequence consists of AU combinations (like [17]). The best performance was obtained by one hidden layer. Although this method can deal with intensity variations, by using continues values for target of feature vectors, it suffers from trapping in local minima due to large number of inputs, i.e. feature vectors. Table 4 shows the facial expression recognition results using JRIP [15] classifier. We used several classifiers such as SVMs for classification of six basic facial expressions, but the results were almost the same. By applying each rule-based classifier we can develop an AU-to-expression converter.

**Table 4.** Facial expression recognition results using JRIP [15] classifier (S=surprise, G=gloomy, F=fear, H=happy, A=angry, D=disgust)

| Confusion matrix for JRIP classifier (total number of samples=2916, correctly classified samples=2675 (91.74%), incorrectly classified samples=241 (8.26%)): | | | | | | | The resulted tree for converting the AU intensities to expressions using JRIP classifier (the value of each AU is between 0 and 1): |
|---|---|---|---|---|---|---|---|
| Classified as → | S | G | F | H | A | D | (AU9 >= 0.268941) and (AU24 == 0) and (AU12 == 0) => class=D (257.0/18.0) |
| S | 581 | 8 | 4 | 1 | 6 | 0 | (AU24 >= 0.5) and (AU23 >= 0.5) => class=A (198.0/0.0) |
| G | 8 | 446 | 0 | 2 | 30 | 0 | (AU1 == 0) and (AU12 == 0) and (AU15 == 0) and (AU20 == 0) and (AU26 <= 0.310026) and (AU27 == 0) |
| F | 21 | 8 | 393 | 1 | 51 | 0 | and (AU7 >= 0.268941) => class=A (30.0/0.0) |
| H | 0 | 0 | 0 | 618 | 0 | 0 | (AU1 == 0) and (AU12 == 0) and (AU15 == 0) and (AU20 == 0) and (AU26 <= 0.083173) and (AU27 == 0) |
| A | 28 | 17 | 0 | 1 | 398 | 18 | and (AU25 >= 0.064969) => class=A (233.0/78.0) |
| D | 6 | 2 | 0 | 1 | 28 | 239 | (AU20 >= 0.5) and (AU4 >= 0.5) => class=F (263.0/0.0) |
| Detailed accuracy by class for JRIP classifier: | | | | | | | (AU20 >= 0.390682) and (AU12 == 0) and (AU27 == 0) => class=F (133.0/0.0) |
| True positive rate | False positive rate | Precision | ROC area | Class | | | (AU23 >= 0.715669) and (AU16 >= 0.715669) => class=F (6.0/0.0) |
| 0.968 | 0.027 | 0.902 | 0.985 | Surprise | | | (AU15 <= 0.017986) and (AU4 <= 0.907687) and (AU5 == 0) => class=G (301.0/1.0) |
| 0.918 | 0.014 | 0.927 | 0.991 | Gloomy | | | (AU4 >= 0.5) and (AU2 == 0) => class=G (162.0/17.0) |
| 0.829 | 0.002 | 0.990 | 0.980 | Fear | | | (AU25 == 0) and (AU1 >= 0.758204) and (AU2 <= 0.847391) => class=G (8.0/0.0) |
| 1.000 | 0.003 | 0.990 | 1.000 | Happy | | | (AU4 >= 0.929) and (AU5 == 0) => class=G (20.0/5.0) |
| 0.861 | 0.047 | 0.776 | 0.958 | Angry | | | (AU27 >= 0.064969) => class=S (456.0/6.0) |
| 0.866 | 0.007 | 0.930 | 0.982 | Disgust | | | (AU12 == 0) and (AU26 >= 0.152609) => class=S (183.0/57.0) |
| | | | | | | | (AU1 >= 0.898605) and (AU24 == 0) => class=S (14.0/1.0) |
| | | | | | | | => class=H (652.0/34.0) |

## 5  Discussion and Conclusions

We proposed an accurate sequence-based system for representation and recognition of single and combinations of facial action units, which is robust to intensity and illumination variations. As an accurate tool, this system can be applied to many areas such as recognition of spontaneous and deliberate facial expressions, multi modal/media human computer interaction and lie detection efforts.

Although the computational cost of the proposed method can be high in the training phase, when the neural network were trained, it needs only some matrix products to reduce the dimensionality of the geometric and appearance features in the test phase. Employing a 3× 3 Gabor kernel and a grid with low number of vertices, we can construct the Gabor representation of the input image sequence and also track the grid in less than two seconds with moderate computing power. As a result, the proposed system is suitable for real-time applications. Future research direction is to consider variations on face pose in the tracking algorithm.

**Acknowledgment.** The authors would like to thank the Robotic Institute of Carnegie Mellon University for allowing us to use their database. Thanks to referees for suggestions and for carefully reading this paper.